\documentclass[conference]{IEEEtran}
\usepackage{cite}
\usepackage{amsmath,amssymb,amsfonts}
\usepackage{algorithmic}
\usepackage{graphicx}
\usepackage{textcomp}
\usepackage{xcolor}
\usepackage{hyperref}
\def\BibTeX{{\rm B\kern-.05em{\sc i\kern-.025em b}\kern-.08em
    T\kern-.1667em\lower.7ex\hbox{E}\kern-.125emX}}
\begin{document}

\title{Embedded Deep Learning for Sleep Staging}

\author{
\IEEEauthorblockN{Engin T\"uretken}
\IEEEauthorblockA{\textit{Embedded Vision System Group}, \textit{CSEM}\\
Neuch\^atel, Switzerland\\
engin.tueretken@csem.ch}
\and
\IEEEauthorblockN{J\'er\^ome Van Zaen}
\IEEEauthorblockA{\textit{Signal Processing Group}, \textit{CSEM}\\
Neuch\^atel, Switzerland\\
jerome.vanzaen@csem.ch}
\and
\IEEEauthorblockN{Ricard Delgado-Gonzalo}
\IEEEauthorblockA{\textit{Embedded Software Group}, \textit{CSEM}\\
Neuch\^atel, Switzerland\\
ricard.delgado@csem.ch}
}

\maketitle

\begin{abstract}
The rapidly-advancing technology of deep learning (DL) into the world of the Internet of Things (IoT) has not fully entered in the fields of m-Health yet. Among the main reasons are the high computational demands of DL algorithms and the inherent resource-limitation of wearable devices. In this paper, we present initial results for two deep learning architectures used to diagnose and analyze sleep patterns, and we compare them with a previously presented hand-crafted algorithm. The algorithms are designed to be reliable for consumer healthcare applications and to be integrated into low-power wearables with limited computational resources.
\end{abstract}

\begin{IEEEkeywords}
CNN, RNN, deep learning, embedded, SoC, sleep, polysomnography, e-health, m-health
\end{IEEEkeywords}

\section{Introduction}
Deep learning (DL) is a branch of machine learning based on artificial neural networks that hierarchically model high-level abstractions in data~\cite{LeCun2015}. Thanks to the recent availability of large-scale labeled datasets and powerful hardware to process them, the field has been successful in many fields including computer vision~\cite{He2016}, natural language processing~\cite{Socher2012}, and speech analysis~\cite{Deng2013}. The technology is likely to be disruptive in many application areas and expected to render conventional machine learning techniques obsolete. Consequently, substantial research is being performed to adopt it in the rapidly-growing wearables and IoT markets~\cite{Miotto2017,VanZaen2019}. According to CCS Insight, the wearable market  is expected to grow from over \$10 billion in 2017 to \$17 billion by 2021.

In this race of building DL-powered wearables, multiple efforts have been made to streamline and simplify current DL frameworks to enable them to be used in the edge. Since wearable devices are at the extreme edge, they need to minimize their computational footprint in order to maximize battery life time for a proper life-style assessment, only the most low-level frameworks, such as CMSIS-NN~\cite{Lai2018}, are simple enough to be considered given the current state of System-On-Chip (SoC) technology.

This paper presents a proof of concept (PoC) using deep learning architectures for sleep staging and evaluates two implementations fully embeddable in low-power SoC. We compare the performance of these architectures with a hand-crafted algorithm designed for low-power variables.

\section{Data}\label{sec:data}
The data for the PoC was taken from the PhysioNet/Computing in Cardiology Challenge 2018 (CinC18)\footnote{\url{https://www.physionet.org/challenge/2018/}}. The dataset was contributed by the Massachusetts General Hospital's (MGH) Computational Clinical Neurophysiology Laboratory, and the Clinical Data Animation Laboratory. The whole dataset includes 1985 subjects who were monitored at an MGH sleep laboratory for the diagnosis of sleep disorders. However, only the training data (994 subjects) was used in this research since it was the only part that was open to the public. The sleep stages of the subjects were annotated by the MGH according to the American Academy of Sleep Medicine~\cite{Berry2012}. Specifically, every 30-second segment was annotated with one of the following labels: wakefulness, rapid eye movement (REM), NREM stage 1, NREM stage 2, and NREM stage 3, where NREM stands for non-REM. For our analysis, we merged NREM stage 1, NREM stage 2, and NREM stage 3. The dataset contained the following physiological signals at a high sampling rate: electroencephalography (EEG), electrooculography (EOG), electromyography (EMG), electrocardiology (ECG), and oxygen saturation (SaO2). We split the dataset into training and test subsets with 90\% for training and 10\% for testing. It is worth noting that this dataset contains a high number of pathological cases since it is a very specific cohort. Therefore, it is not representative of the general population but constitutes a challenging medical-grade dataset.

\section{Methods}\label{sec:algorithms}
\subsection{Hand-designed}
\begin{figure}
    \centering
    \includegraphics[width=\columnwidth]{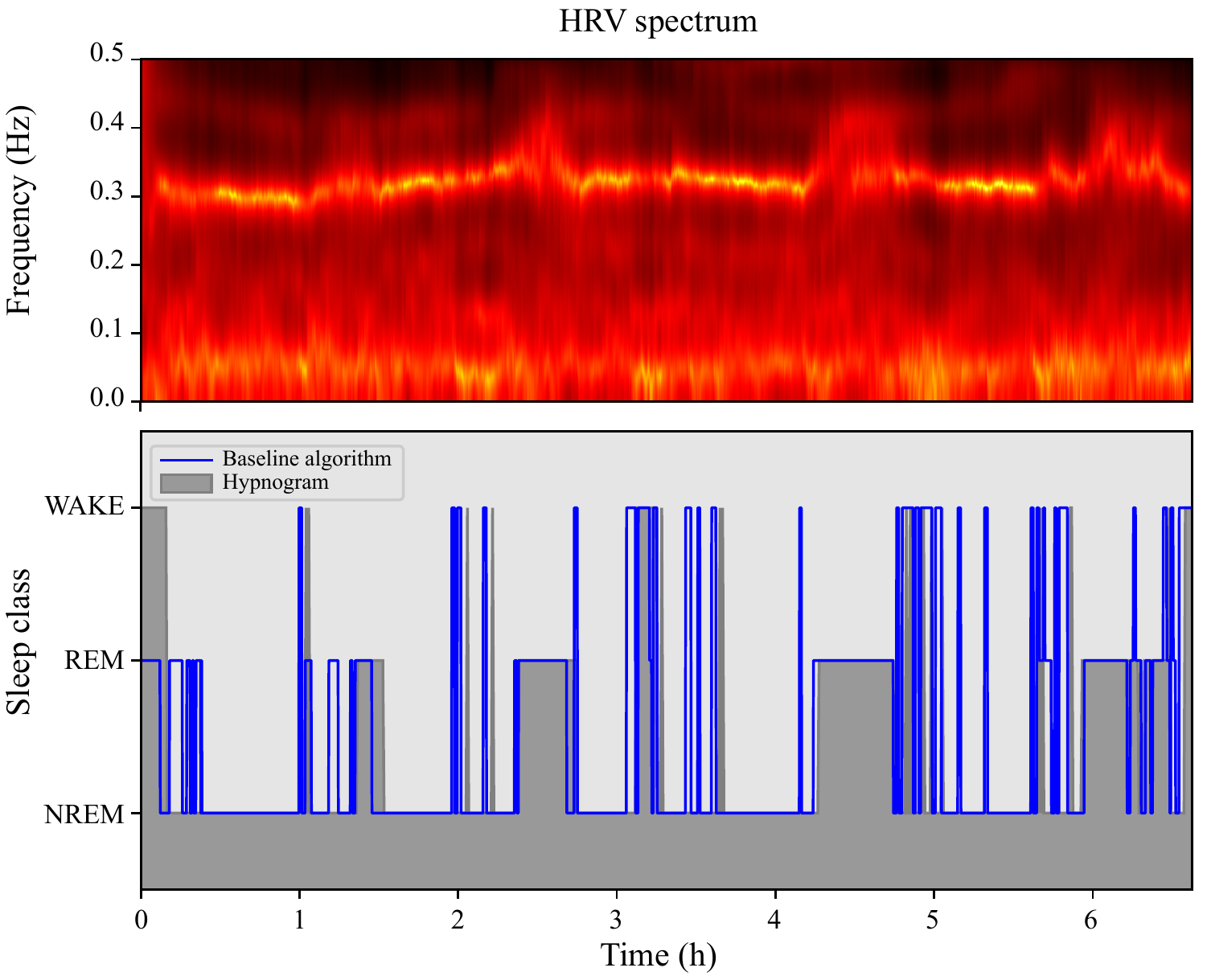}
    \caption{\label{fig:hrv}Time frequency representation of the HRV spectrum (top) and estimated hypnogram versus reference (bottom).}
\end{figure}

In order to establish a baseline, we used our previously-descried algorithm based on physiological cardiorespiratory cues~\cite{Renevey2017} extracted from photoplethysmographic sensors~\cite{Renevey2018}. For a healthy population, this algorithm proved to achieve a sensitivity and specificity for REM of 89.2\% and 77.9\% respectively; and for NREM 83.4\% and 84.9\% respectively. The algorithm was exclusively based on the analysis of the heart rate variability (HRV) and movement (Fig.~\ref{fig:hrv}).

\subsection{Deep learning}
The input to the algorithms is a short temporal sequence (around 8.5 minutes) of HRV values at 4Hz and a binary value that denotes whether the subject moved. We developed several deep learning architectures for sleep staging and evaluated two embedded implementations:
\begin{itemize}
\item Multi-Layer Recurrent Neural Networks (ML-RNNs): RNNs are powerful architectures designed to model long-term temporal relations in the data. They have been shown to work very well in natural language processing.
\item Convolutional Neural Networks (CNNs): CNNs are neural networks which can capture local patterns with increasing semantic complexity in spatiotemporal data.
\end{itemize}

In our implementation, the CNN and the ML-RNN architectures employ fully connected and normalization layers, and the latter is based on an efficient version of LSTM with forget gates. More specifically, the ML-RNN architecture is comprised of two LSTM layers respectively with 128 and 32 hidden units, which are followed by two fully connected layers with bias. The CNN architecture contains eleven convolutional layers with a temporal size of 3, a single max-pooling layer, and two fully connected layers (see Table~\ref{tb:cnn}). The convolutional layers are followed by batch normalization and ReLU units.
\begin{center}
\begin{table}{
\caption{\label{tb:cnn}Layers of the CNN architecture.}}
\hfill{}
\begin{tabular}{c|ccc}
\hline\hline
Layer type & Stride & Filter shape & Input size \\
\hline
Conv & 1 & 3 $\times$ 3 $\times$ 32 & 2048 $\times$ 3\\
Max Pool & 1 & 2 $\times$ 1 & 2048 $\times$ 32\\
Conv & 2 & 3 $\times$ 32 $\times$ 32 & 1024 $\times$ 32\\
Conv & 2 & 3 $\times$ 32 $\times$ 48 & 512 $\times$ 32\\
Conv & 2 & 3 $\times$ 48 $\times$ 64 & 256 $\times$ 48\\
Conv & 2 & 3 $\times$ 64 $\times$ 64 & 128 $\times$ 64\\
Conv & 2 & 3 $\times$ 64 $\times$ 64 & 64 $\times$ 64\\
Conv & 2 & 3 $\times$ 64 $\times$ 64 & 32 $\times$ 64\\
Conv & 2 & 3 $\times$ 64 $\times$ 64 & 16 $\times$ 64\\
Conv & 2 & 3 $\times$ 64 $\times$ 64 & 8 $\times$ 64\\
Conv & 1 & 3 $\times$ 64 $\times$ 64 & 4 $\times$ 64\\
Conv & 1 & 1 $\times$ 64 $\times$ 64 & 4 $\times$ 64\\
FC & 1 & 256 $\times$ 256 & 1 $\times$ 256\\
FC & 1 & 256 $\times$ 21 & 1 $\times$ 256\\
\hline\hline
\end{tabular}
\hfill{}
\end{table}
\end{center}

\section{Results}\label{sec:results}
The RNN-based network brings approximately 20\% improvement in mean accuracy over the baseline method and the CNN approximately 35\% (see Table~\ref{tb:accuracy}). Higher accuracy obtained by the CNN network suggests that high frequency patterns in the data, which RNNs are ineffective at modeling, is informative for the sleep staging task. As shown in Table~\ref{tb:accuracy}, both architectures are small in size and require limited processing resources to run. Their particular execution time will depend on the SoC that is selected.
\begin{center}
\begin{table}{
\caption{\label{tb:accuracy}Accuracy and complexity of the proposed algorithms.}}
\hfill{}
\begin{tabular}{c|ccc}
\hline\hline
Algorithm & Accuracy & MACs & Parameters \\
\hline
Hand-designed (baseline) & 47\% & \textemdash & \textemdash \\
ML-RNN & 68\% & 1.2M & 1.2M\\
CNN & 81\%  & 6.2M & 166K \\
\hline\hline
\end{tabular}
\hfill{}
\end{table}
\end{center}

\section{Conclusion}\label{sec:conclusion}
We presented results for two prominent DL architectures used to diagnose and analyze sleep patterns on the CinC2018 dataset, and compared them to a hand-crafted algorithm. The DL architectures outperform by more than 20\% the hard-designed baseline which was developed on a database on healthy subjects. Inter-observer agreement of expert sleep scorers on more reliable and multi-modal sensory data such as EEG and EOG is less than 85\%~\cite{Norman2000,Penzel2013}. Considering that we used only HRV as input and the pathological nature of the dataset used to train the efficient CNN model, an average accuracy of 81\% on 3-class sleep staging is very promising. Future work includes embedding the networks on a resource-limited processor from the ARM Cortex-M family or a low-power neural network hardware accelerator, such as the GAP8.

\bibliography{IEEEabrv,references}

\end{document}